\documentclass[a4paper,twocolumn,10pt]{article}
\usepackage{eurosis}
\usepackage{url}
\usepackage{natbib}

\usepackage{graphicx}
\usepackage{gensymb}

\bibpunct[; ]{(}{)}{,}{a}{}{;}

\begin{document}

% vmi szimulacios specifikus dolgot meg beleirni par mondatot, pl h egyik valos ideju a amsik nem, a nem valos ideju meg akkora terkeppel meg forgalpommal nem tud boldogulni, amaz meg nem tud robotkarokat modellezni
% van industry 4.0 track is
\title{FATHER: FActory on THE Road}

%hogyan teszek bele tobb author-t?
\author{G{\'e}za Szab{\'o} \\
Ericsson Research, Budapest, Hungary \\
email: \texttt{geza.szabo@ericsson.com}
\and Bal{\'a}zs T{\'a}rnok, Levente Vajda, J{\'o}zsef Pet\H{o}, Attila Vid{\'a}cs \\
HSN Lab, Department of Telecommunications and Media Informatics,\\ Faculty of Electrical Engineering and Informatics,\\ Budapest University of Technology and Economics, Hungary \\
email: \texttt{vidacs.attila@vik.bme.hu}}

\date{}

\maketitle

\thispagestyle{empty}

\keywords{multi-simulation environment, robotic cell, mobile communication, C-ITS, cyber-physical-system}

\begin{abstract}
In most factories today the robotic cells are deployed on well enforced bases to avoid any external impact on the accuracy of production.
In contrast to that, we evaluate a futuristic concept where the whole robotic cell could work in a moving platform. Imagine a trailer of a truck moving along the motorway while exposed to heavy physical impacts due to maneuvering. The key question here is how the robotic cell behaves and how the productivity is affected. We propose a system architecture (FATHER) and show some solutions including network related information and artificial intelligence to make the proposed futuristic concept feasible to implement.
%To make the evaluation of the concept possible we connected three simulators together: one that is capable of simulating the traffic on the road, one that is a network simulator that simulates the V2X network and a rigid-body simulator that makes detailed simulation on moving robot cell. We discuss how the simulators are connected and affect each other to get detailed productivity information on the robot cell performance.  
\end{abstract}

\section{INTRODUCTION}\label{intro}
In existing factories, the production lines including robot cells are designed and optimized for specific tasks.
%In existing factories, the production lines are designed for producing specific products. A robot cell is designed and optimized for its specific task. 
These lines are typically used for month or even for years without reconfiguration. Devices (e.g., robot arms) are precisely installed on well enforced bases to avoid any external impacts, and to achieve the required pose accuracy. This precise installation of robotic equipment is time consuming. 
%The emerging Machine-as-a-Service (MaaS) scenarios induce the usage of robotic cells that are on-demand deployed and transportable. In this concept the long-lasting installation procedure and the rigid static environment are costly and thus clearly undesirable. 
To add to this, we take a huge leap forward and attempt to introduce and analyze use-cases where the whole robotic cell works in a moving environment. For example, the application of this concept would enable a robotic cell to be deployed on the go on a truck, train, airplane or on a ship. This daring ‘FActory on THE Road’ (FATHER) scenario would mean a radical change compared to recent manufacturing environments. It immediately raises many new challenges and problems to solve on one hand, but there would be many benefits as well. %Plenty of transportation vehicles cross the roads day-by-day delivering raw materials or finished products. If a robotic cell could work properly on a moving vehicle, it would save time and could result in growth in production. Just think of the autonomous yard maneuvering use-cases where unloading could be sped-up by preparing stuff to unload before and during parking.

Our goal is to show that FATHER is feasible, by analyzing a robotic cell on the move that is exposed to external physical forces in a simulation environment. 
We propose techniques to overcome the difficulties with the help of Cooperative-Intelligent Transport Systems (C-ITS).
We focus on a special use case for C-ITS in connection with manufacturing.

\section{ROBOT CELL ON THE MOVE}\label{sec:modelling}
%----------------------------------

Our main goal is to show how a robotic cell can withstand external forces occurring on the move. To achieve this goal, we take the Agile Robotics for Industrial Automation Competition (ARIAC) 2018 environment \citep{ariac2018} as a baseline, and extend it to serve our needs. First, we modified the static environment and mobilized it. The next step was to apply external forces from different sources to the modified model. Our final goal is to examine the productivity changes in the moving system, and based on the results, propose suggestions to decrease the impact of the external forces.

\begin{figure*}
\centering
\includegraphics[width=.99\textwidth]{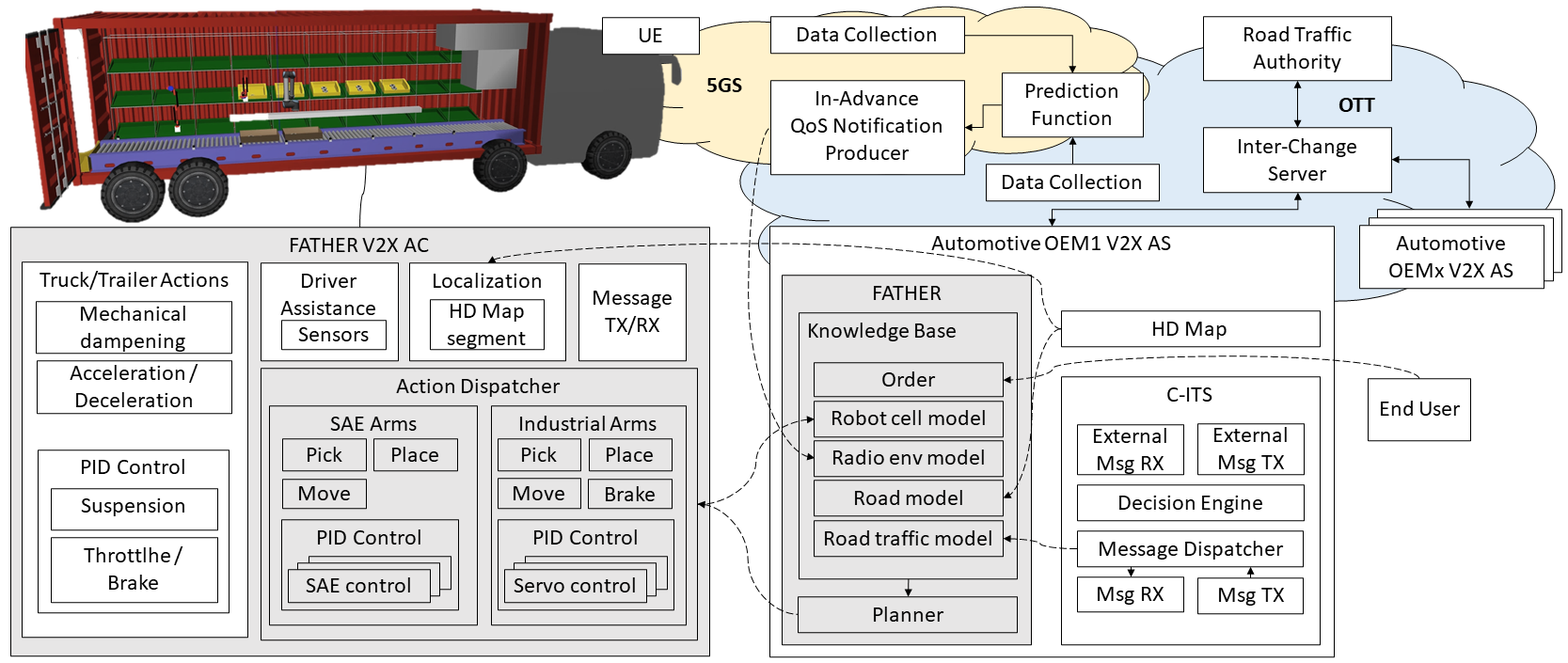}
\caption{The architecture of the proposed systems}
\label{fig:system}
\end{figure*}

\subsection{Putting the robot cell on wheels}

Our intention was to apply acceleration data sets programmatically on the moving robot cell. 
Since Gazebo is able to apply forces on the root model, a plugin was implemented that controls the model dynamics.

%\subsubsection{Root model and Gazebo plugin}
%A root model was implemented that would move when an external force is applied to it. 
%This model was designed to block the movement of some models inside the shipping container while allowing others to move freely. We fixed the yellow storage bins and the conveyor belt relative to the shipping container. We also created fixed joints between the shipping container and the base of the robotic arms. On the other hand, the product parts and the shipping boxes could move freely and could collide with other objects.

%The \texttt{AddForceAtRelativePosition()} function adds a force in world frame coordinates to the body at a position relative to the center of mass which is expressed in the link's own frame of reference. To get the force attributes of the function we subscribed to our \texttt{/imu} topic within the plugin. 
%
%\subsubsection{Acquiring IMU data}
%We published acceleration values to the \texttt{/imu} topic obtained from various sources. 

We obtained acceleration values from various sources.
Firstly, we recorded real acceleration data during a city bus ride with an Android phone's IMU and replayed that to the simulator
%\texttt{/imu} topic 
so our plugin could use these values to add the force to the shipping container. 

Secondly, we extracted acceleration data from the
Euro Truck Simulator~2~(ETS2) \citep{ets2}. ETS2 is a truck simulator game in which the player can choose a truck, pick up and deliver cargo from/to various locations. It is also possible to modify the vanilla game. 
%A mod is an alteration that changes one or more aspects of a video game, such as how it looks or behaves. 
The ETS2 game engine contains the Telemetry SDK %~\citep{scssdk} 
that provides access to the telemetry of the player’s vehicle for third-party applications. There are telemetry fields on the acceleration of the truck and the trailer as well.
\section{ARCHITECTURE OF FATHER}\label{sec:architecture}
%---------------------------------

Figure~\ref{fig:system} shows the architecture of the proposed system.
The general architecture consists of the V2X Application Client (V2X-AC), the V2X Application Server (V2X-AS), as well as the Inter-Change Server as an optional component ensuring the interoperability of C-ITS services across different V2X application servers \citep{v2xericsson}.
The 5G System (5GS) provides the radio link, the network slices and data collection that are required for the prediction function. %5G mobile networks will be the key to provide connectivity for vehicle to vehicle (V2V) and vehicle to infrastructure (V2I) communications. Over the Top (OTT) services in this specific context hosts the Automotive Original Equipment Manufacturer (OEM) V2X-AS--V2X-ACs and provides data for the prediction function while communicates with the Road Traffic Authority and other Automotive OEMs V2X-ASs.
%DONE TBD [Geza]: 5GS, OTT-krol irni meg, prediction function

%\subsection{V2X application server}
%V2X-AS is at the backend or edge servers that are accessible by V2X-ACs via cellular networks. %A V2X-AS may include Message Reception (Msg RX) and Message Transmission (Msg TX), which handle the message communication with V2X-AC, and a Decision Engine, which processes the received message and makes decision on its further dissemination to V2X-ACs. The Message Dispatcher, if present, disseminates the message to the intended receivers according to the requirements of the services.

%V2X-AS may be implemented at the backend system of different stakeholders such as automotive Original Equipment Manufacturers (OEMs), third-party ITS service providers, or Road Traffic Authorities (RTAs). For V2X-ACs that are connected to different backend systems to support interoperable C-ITS and Advanced Driver-Assistance System (ADAS) services, V2X-ASs need to be interconnected via the external TX and RX interfaces for backend systems. In this general architecture, the inter-change server interconnects different backend systems via standard interfaces.

In our proposed architecture, the V2X-AS hosts the %\emph{HD Map} \citep{hdmaps} functionality and 
the FATHER application backend. % as well.
The FATHER backend consists of a \emph{Planner}, and a \emph{Knowledge Base} that feeds the Planner with input.

\subsubsection{Planner}\label{sec:planner}
The planner is an AI tool with a main goal to produce valid plans from a starting state to a specified goal state with given actions. The cost function that the planner tries to minimize is the overall duration of the whole plan. 
Our baseline is based on \citep{qocplanning} in which modular arms are introduced and utilized in parallel with an industrial arm to speed up a series of pick and place tasks. %...where the modular robot arms are introduced.

All the above information coming from various sources like the \emph{Road model}, \emph{Road traffic model} and \emph{Radio environment model} can be translated into logical predicates and considered in the planner.

% \subsection{V2X application client}\label{sec:v2xapp}
% The V2X-AC can be at vehicles, personal devices, or road-side units which are all provisioned with cellular connectivity. Function-wise, V2X-AC is divided into V2X-AC Transmission (TX) and V2X-AC Reception (RX).
% The V2X-AC TX transmits messages to V2X-AS using cellular uplink unicast communication. The V2X-AS uses cellular downlink unicast, multicast or broadcast to transmit messages to V2X-ACs (RX).
% We use the traffic wave mitigating method to reduce external forces on the robot cell. This can be considered in the Planner.

% \subsubsection{Action dispatcher}

% The planner is connected to the action dispatcher to realize the planned actions within the environment. 
% The action dispatcher has various actions for the conveyor belt, the industrial robot arm and the SEA arms (high level actions). The action dispatcher calls the device specific PID controllers module by module as all the devices are controlled in joint space (low level actions). In case of arriving at the goal position the controllers call the action feedback which setups the specific predicates in the Knowledge Base. Other actions in connection with the truck or trailer can be called from the action dispatcher as well, for example, mechanical dampening or active suspension, etc.

%---------------------------
\section{ASSISTING THE ROBOT CELL TO WITHSTAND EXTERNAL FORCES}\label{sec:discussion_plans}

\begin{figure*}
\centering
\includegraphics[width=.79\textwidth]{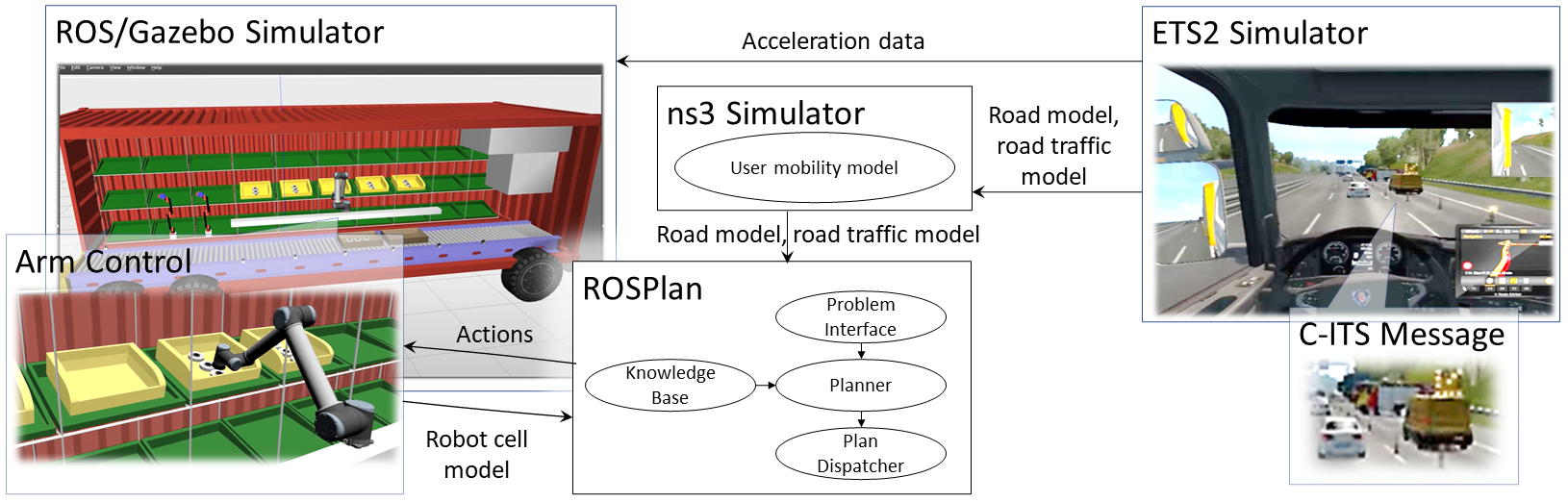}
\caption{The overview of the working mechanism of the proposed system.}
\label{fig:general_working}
\end{figure*}

%\subsection{Applying C-ITS}\label{sec:plannerros}
Application of network provided information enables the use of more sophisticated mechanism to decrease the impact on the productivity of the robot cell even more.

All the information coming from the C-ITS, like the Road, Road traffic and Radio environment models can be translated into \emph{timed initial literals} into the planner.

Timed initial literals add to the temporal part of Planning Domain Definition Language (PDDL) 2.1 (level 3) a way of defining literals that will become true at a certain time point, independently of the actions taken by the planner. A typical use is to formulate time windows and/or goal deadlines~\citep{DBLP:journals/corr/abs-1109-5663}.

%\subsubsection{Synchronization with ETS2}
Figure~\ref{fig:general_working} shows the working mechanism of the integrated simulation systems.
A video on the working mechanism can be seen on \citep{fatherdemovideo}.
ETS2 version 1.31 introduces the random road events feature in the game. This feature can trigger incidents on the motorway. Upcoming road events are usually indicated by a sign in the game. We used the visual feedback of these signs to send the C-ITS message to the planner that in the upcoming 30 sec may contain directional changes or sudden deceleration. 

% uj resz az ns3-al
For the proposed C-ITS based method we applied ns3~\cite{ns3}, a discrete-event network simulator to apply the effect of C-ITS data propagation between the robot cell and the C-ITS system. Ns3 is applied with mmWave ns-3 module \cite{8344116} in tap bridge mode to simulate the effects of 5G. The mobility model parameters are synchronized with the ETS2 road traffic parameters and the velocity of the truck.

C-ITS messages were taken into account for the action dispatching of the industrial arm. A new condition was introduced in the \emph{domain specification} of the industrial arm's pick and place actions. It requires that the \texttt{til\_enable} predicate to be true at the beginning of the action. 
%Figure~\ref{figure:plans_cits} shows the resulting plan. It can be seen that the SEA arms were used during the first road event, but the industrial arm was not. During the second road event not even the SEA arms were used. 
%Most likely, there was a dependency on using the SEAs with the industrial arm.
%The total duration of the plan is shorter than in the case when all arms are put on hold for the duration of the road event, as the level of parallel execution is higher in this case.

% \subsubsection{Other use cases} 
% The core mechanism required to build useful plans is to keep the knowledge base up to date with the available information sources. As an example, the road model can give information on upcoming bumps on the road that can trigger lane change for the truck or simply prepare the active suspension system.
% Other use cases that can be similarly implemented include the application of the QoS prediction data to schedule downloading of the HD maps, the robot cell plans and the point clouds for object recognition and grab pose detection. 

%--------------------------------------------------------
\section{EVALUATION}\label{sec:evaluation}
%--------------------------------------------------------
%\subsection{ARIAC scoring}
For evaluation purposes we use the ARIAC~\citep{ariac2018} scoring as KPIs for the different setups. %These are not typical robotic related KPIs like accuracy or repeatability, but they provide a normalized indicator of the performance of the robot cell during a complex scenario.
%The considered two KPIs were as follows: The \emph{Total Game Scor}e (TGS) is a compound of the presence of required parts, correct position and orientation, and bonus points when all parts are present. The \emph{Total Processing Time} (TPT) is duration between the time the first order is issued until the end of the trial.

% We tested the system with various setups. The order to be fulfilled consisted of two identical shipments, each containing 6 product parts. That is, altogether 12 part pieces were to be collected an placed into two shipping boxes.

\begin{figure}
\centering
\includegraphics[width=.45\textwidth]{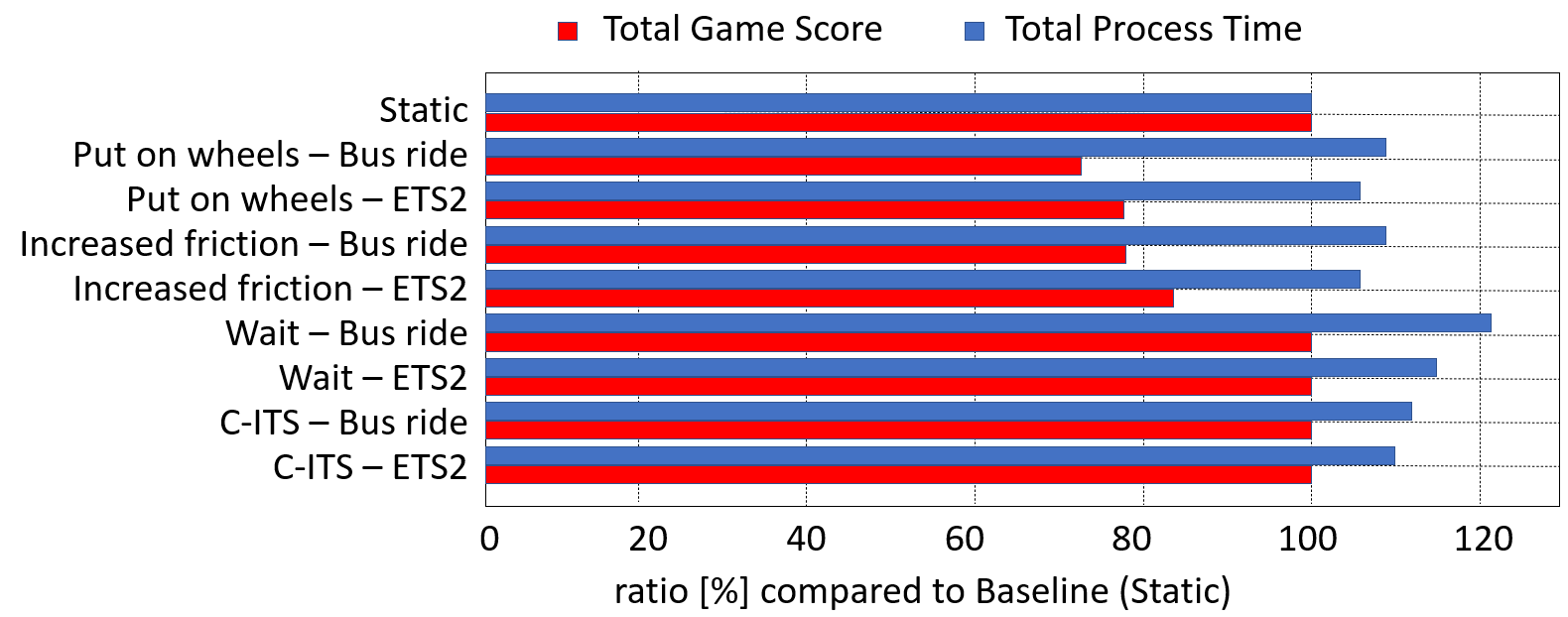}
\caption{ARIAC scores.}
\label{fig:ariac_scores}
\end{figure}

Figure~\ref{fig:ariac_scores} summarizes the ARIAC scores for the measurements, that can be explained as follows.

\subsubsection{Static}
The \emph{Static} system was chosen as a baseline where no mobility effects were present at all, the robot cell was still. It achieved 36 total points as final score %\citep{ariac_scoring}
\citep{ariac2018}. %This 36 is the achievable maximum, since 1 point was awarded for each requested and delivered part, 1 more point when the part was placed precisely in the box with right orientation, and an extra bonus point was given if all requested parts were present in the shipment. 
All the other cases were compared to this result.

\subsubsection{Put it on wheels}
We evaluated the case when the whole robot cell was carried by the trailer, but no extra defense mechanisms were applied to fend off the external impacts. Two acceleration data sets were applied coming from 1) a real bus ride, and 2) the ETS2 simulation where one emergency brake occurred after a road event.

\subsubsection{Replanning with TIL}
Applying C-ITS data in the planner resulted in reduced TPT compared to the previous Wait case as the planner can increase the level of parallel execution of actions with the robot arms while still maintaining the 100\% TGS.
These results further motivate the application of C-ITS. %We believe that other -- even seemingly extreme -- manufacturing related use cases can be supported in the long run. The benefits of communication can reach out to the physical world and eliminate the impacts of sudden movements in a factory-on-the-road environment, providing similar production KPIs as a stationary setup.

%It is also important to note that there is no disadvantage or disincentive of the application of C-ITS. For various safety reasons and optimization of logistics, communication will be a core component in today's and future trucks.

%------------------------------------------
\section{CONCLUSIONS}\label{sec:conclusion}
%------------------------------------------
Through evolution in automation and cellular V2X capabilities, the automotive world is undergoing a transformation which will increase safety, efficiency and mile-economy.
In this paper we took a huge leap forward compared to the previously mentioned common use cases and introduced the ‘FActory on THE Road’ concept in which the whole robotic cell works in a moving environment. 
%The proposed FATHER architecture integrates C-ITS messages into a robot cell use case which is operated and moved along the motorway to minimize idle time of MaaS scenarios while also establish the possibilities of pushing just-in-time manufacturing to the extremes.
%The moving robot cell induces some issues in connection with the external forces on the moving robot cell due to the acceleration of the truck.
%We dealt with these with various techniques including mechanical, communication and AI related ones. We evaluated the proposed scenarios in a physical simulator to check how the robot cell productivity KPIs are altered compared to the static deployed case.
The evaluation shows that with the application of C-ITS related information the system can achieve the same productivity accuracy as the static version while the execution time is only increased with 10\%.
%uj simulation related resz
We also showed how a real-time game engine, a non-real time rigid body simulator can be connected with a discrete event network simulator in bridge mode.
It is quite possible that we will see the benefits of some similar deployments in the near future!

\bibliography{refs}

\end{document}